\documentclass{article}
\usepackage{spconf,amsmath}
\usepackage{booktabs}
\usepackage{enumitem}
\usepackage{multirow}
\usepackage{bigstrut}
\usepackage{graphicx}
\usepackage[numbers,sort]{natbib}
\usepackage{amsfonts}
\usepackage{subcaption}
\let\OLDthebibliography\thebibliography
\renewcommand\thebibliography[1]{
  \OLDthebibliography{#1}
  \setlength{\parskip}{0pt}
  \setlength{\itemsep}{0pt plus 0.3ex}
}

\begin{document}\sloppy
\topmargin=0mm
\def\x{{\mathbf x}}
\def\L{{\cal L}}

\title{Disentangle and Denoise: Tackling Context Misalignment for Video Moment Retrieval}

\name{
 \parbox{\linewidth}{\centering
Kaijing Ma$^1$$^2$$^*$\thanks{$*$ This work was done during an internship at TeleAI.}, Han Fang$^2$, Xianghao Zang$^2$, Chao Ban$^2$, Lanxiang Zhou$^2$, Zhongjiang He$^2$ \\
Yongxiang Li$^2$, Hao Sun$^2$, Zerun Feng$^2$, Xingsong Hou$^1$$^{\dagger}$\thanks{$\dagger$ Corresponding author.}}
}
\address{$^1$ Xi'an Jiaotong University $^2$ TeleAI\\
  }

\maketitle

\begin{abstract}

Video Moment Retrieval, which aims to locate in-context video moments according to a natural language query, is an essential task for cross-modal grounding.
Existing methods focus on enhancing the cross-modal interactions between all moments and the textual description for video understanding. 
However, constantly interacting with all locations is unreasonable because of uneven semantic distribution across the timeline and noisy visual backgrounds.
This paper proposes a cross-modal Context Denoising Network (CDNet) for accurate moment retrieval by disentangling complex correlations and denoising irrelevant dynamics.
Specifically, we propose a query-guided semantic disentanglement (QSD) to decouple video moments by estimating alignment levels according to the global and fine-grained correlation.
A Context-aware Dynamic Denoisement (CDD) is proposed to enhance understanding of aligned spatial-temporal details by learning a group of query-relevant offsets. Extensive experiments on public benchmarks demonstrate that the proposed CDNet achieves state-of-the-art performances.

\end{abstract}

\section{Introduction}
\label{sec:intro}

\begin{figure}[t]
    \centering
    \includegraphics[width=0.5\textwidth]{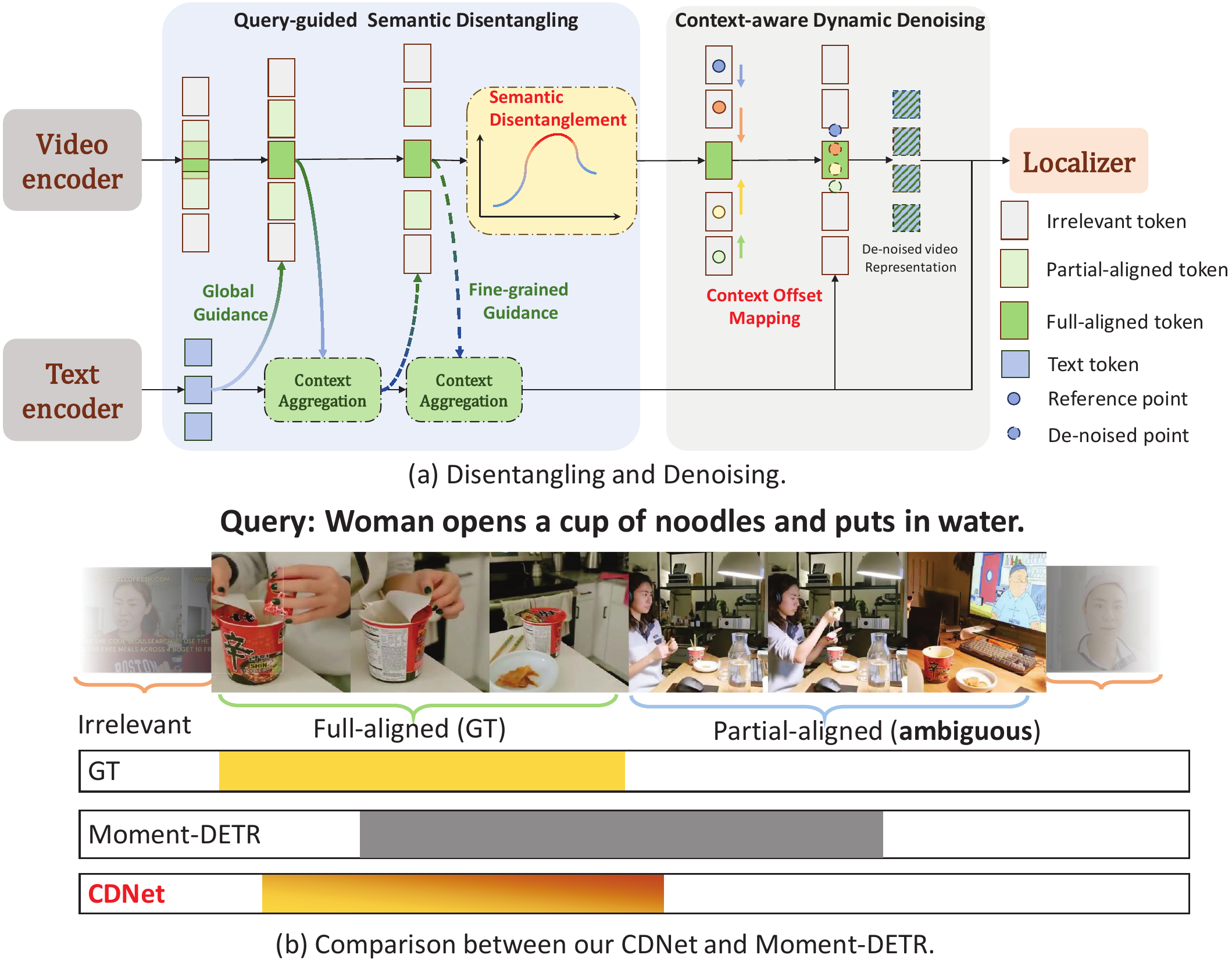} 
    \caption{
    (a) Our proposed disentangle and denoising method. (b) Results comparison when ambiguity occurs in video. } 
    \label{fig:figure1} 
\end{figure}

Videos are gradually becoming the primary carrier of contemporary streaming information. 
Retrieving content in noisy internet videos is increasingly becoming a growing demand. Video Moment Retrieval (VMR), which aims to locate moments described by a given natural language query in a video, faces challenges due to the inherent difficulty of performing structured segmentation of video semantics. 


Inspired by some existing works in image domains~\cite{dosovitskiy2020vit,carion2020end}, current methods~\cite{NEURIPS2021_62e09734,moon2023query,lin2023univtg} commonly leverage Transformers~\cite{Vaswani_Shazeer_Parmar_Uszkoreit_Jones_Gomez_Kaiser_Polosukhin_2017} for global temporal understanding and multi-modal perception with modalities coupled.
They overlook that limited textual description is only temporally aligned with part of a video in the VMR task.
Resultantly, visual context with similar semantics can interfere with the localization of target moments. Taking an example from the QVHighlights~\cite{NEURIPS2021_62e09734} dataset (e.g., Fig.~\ref{fig:figure1} (b)). Most entities described in the text query appeared in video segments outside of the ground truth. Such semantically similar moments are particularly confusing when using global modeling approaches directly.

Furthermore, we find that even for temporally well-aligned multi-modal retrieval tasks, such as text-video-retrieval~\cite{han2022temporal}, there is still much redundancy, repetition~\cite{centerclip}, and noisy spatio-temporal information~\cite{xia2022cross} in video clips. Such semantic noise and redundancy lead the model to be unable to capture fine-grained key context content~\cite{liu2022learning}, such as subtle motions, thus yielding sub-optimal results.


An intuitive approach to prevent semantically similar moments from hindering the localization of target boundaries is first to decouple the correlation of video and text and examine their alignment level.
We argue that the alignment between video sequences and query in the VMR task can be categorized into multiple levels: \textbf{irrelevant}, \textbf{partial-aligned}, and \textbf{full-aligned}. 
Unlike previous approaches, we advocate for disentangling correlations between modalities before global modeling. Specifically, we introduce Query-guided Semantic Disentangling (QSD) based on dual-contrastive learning. This module is designed to disentangle the visual-textual relevance at two scales: global and fine-grained. Benefiting from the disentangling process, we explicitly model the alignment levels between modalities via a relevance score.

Subsequently, inspired by Deformable DETR~\cite{zhu2020deformable}, we introduce Context-aware Dynamic Denoising (CDD).
CDD incorporates a learnable position offset based on multi-modal perception to re-sample visual features. It compels the model to focus on semantically relevant fine-grained context by employing a corrected multi-head attention mechanism.

Our key contributions are: \textit{\textbf{(i)}} We propose Query-guided Semantic Disentangling~\textbf{(QSD)} network based on dual-level contrastive learning. This facilitates separating visual features from irrelevant or partial-aligned elements, enhancing the representation of full-aligned visual information. \textit{\textbf{(ii)}}~We introduce a Context-aware Dynamic Denoising \textbf{(CDD)} network, enabling the model to focus on fine-grained key context rather than noisy backgrounds. \textit{\textbf{(iii)}}~We achieve state-of-the-art performance on QVHighlights benchmark and competitive results on Charades-STA~\cite{gao2017tall} and TACoS~\cite{regneri2013grounding}. 

\section{Related Work}
\textbf{Video Moment Retrieval} is initially proposed by CTRL~\cite{gao2017tall}. Early approaches mainly employ a two-stage approach~\cite{gao2017tall,Ge_Gao_Chen_Nevatia_2018,liu2018cross,liu2018attentive}. They sample moments as candidates first, and rank them after, which is computationally inefficient. Subsequent research focus on efficient multi-modal interactions, introducing proposal-based~\cite{Zhang_Peng_Fu_Luo_2019,Zhang_Peng_Fu_Lu_Luo_2022} and proposal-free methods~\cite{Yuan_Mei_Zhu_2019,Zhang_Sun_Jing_Zhou_2020}.
Recently inspired by the success of Transformer~\cite{Vaswani_Shazeer_Parmar_Uszkoreit_Jones_Gomez_Kaiser_Polosukhin_2017} in various image tasks~\cite{carion2020end,dosovitskiy2020vit}, Moment-DETR~\cite{NEURIPS2021_62e09734} emerges. It adapts DETR for VMR and introduces Query-based Highlight Detection. The latter task necessitates saliency scoring of video content based on detected moments, imposing higher demands for nuanced semantic understanding.
However, Moment-DETR and its successive works~\cite{moon2023query,liu2022umt} excessively rely on the Transformer's robust modeling capabilities, disregarding the impact of the attention mechanism's lack of inductive biases~\cite{dosovitskiy2020vit,Xu_Zhang_Zhang_Tao_2021} on visual tasks. Feed concatenated video and text tokens into self-attention~\cite{NEURIPS2021_62e09734}, or combining them via cross-attention~\cite{moon2023query}, may contradict the inductive biases associated with temporal sequence. We advocate for enhancing multi-modal perception abilities by feeding features, decoupled after correlating, into meticulously designed Transformer-improved models. 

\textbf{Video Text Alignment.} Aligning videos and text is crucial in understanding motion and temporal coherence. In particular, the lack of fine-grained modality interaction makes it harder to align clips and captions temporally.
Some works have relied on attention mechanisms~\cite{Vaswani_Shazeer_Parmar_Uszkoreit_Jones_Gomez_Kaiser_Polosukhin_2017} to extract key information from videos~\cite{Yu_Ko_Choi_Kim_2016}, while others preserve visual information by composing pairwise joint representation using 3D spatio-temporal features~\cite{Yu_Kim_Kim_2018}. The HGR~\cite{chen2020fine} model decomposes video-text pairs into global-to-local levels. Thus far, how these task-specific architectures can be integrated into VMR's video text alignment has not been studied. In this paper, we are the first to propose a simple yet effective token-aware contrastive loss for fine-grained video-text alignment.

\begin{figure*}[t]
    \centering
    \includegraphics[width=1\linewidth]{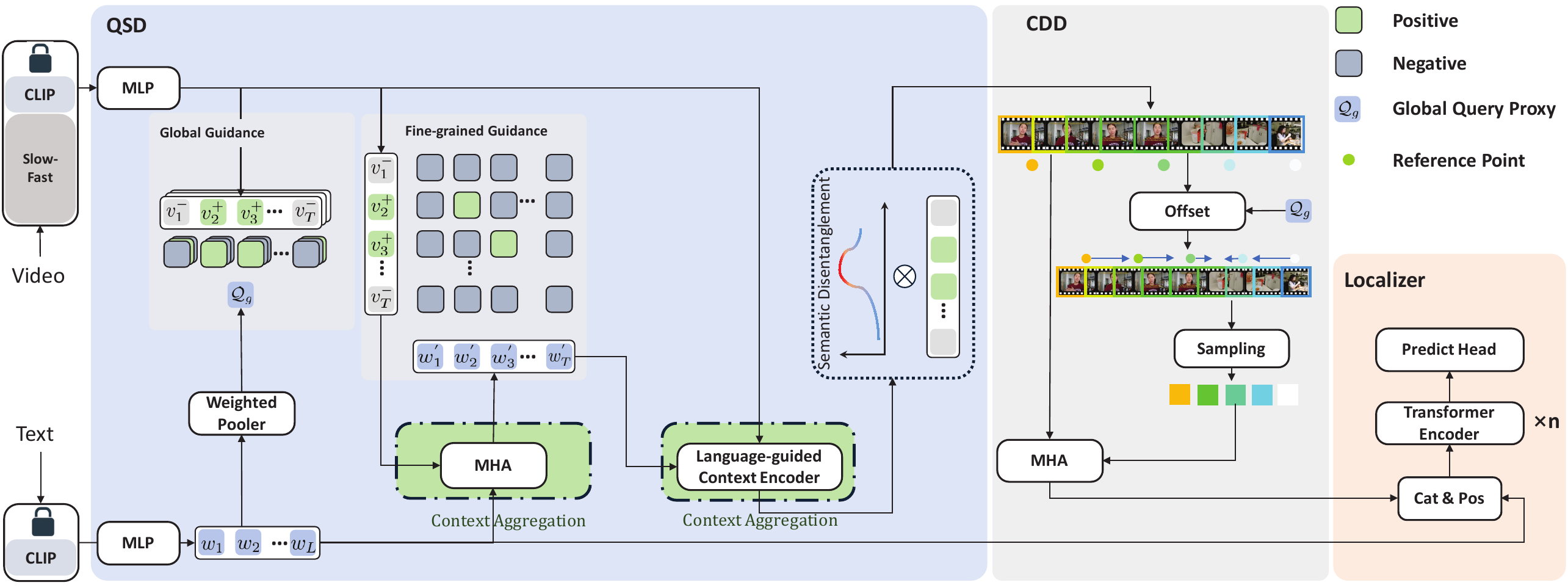} 
    \caption{Illustration of CDNet. (1) In Query-guided Semantic Disentangling (QSD), we use a dual-stream contrastive learning approach to disentangle the visual and textual correlations separately by Global and Fine-grained Guidance. (2) Subsequently, visual and textual features are fed into the Context-aware Dynamic Denoising(CDD) for refined temporal and multi-modal context comprehension. MHA denotes multi-head attention, CMHA signifies corrected multi-head attention, and Offset represents a lightweight convolutional network to guide the model in visual feature re-sampling. (3) The concatenated features are then fed into the Localizer for the ultimate localization.} 
    \label{fig:overall} 
\end{figure*}
\section{Method}

As illustrated in Fig.~\ref{fig:overall}, we first adopt the pre-trained encoder to encode video clips and text. Following Moment-DETR \cite{NEURIPS2021_62e09734}, 
we divide the video into $T$ clips, each representing a short duration of the video. We adopt SlowFast~\cite{feichtenhofer2019slowfast} and CLIP~\cite{radford2021learning} to encode each video clip into video representation $\mathcal{V}=\{v_1,v_2,...,v_T\}$. Then we adopt the text encoder of CLIP~\cite{radford2021learning} to embed every word as textual representation: $\mathcal{Q}=\{w_1,w_2,...,w_L\}$, where $L$ is the number of words.
To align the video representation $\mathcal{V}$ and textual representation $\mathcal{Q}$ into the same dimensional space, two linear projections (MLP) are employed to encode them into $d$-dimensional space, denoted as $\mathcal{V}^d$ and $\mathcal{Q}^d$.

\subsection{Query-guided Semantic Disentanglement}
\label{subsec:saliency}

We propose Query-guided Semantic Disentanglement (QSD) to resolve semantic ambiguity arising from similar video segments. There are three modules: global guidance, fine-grained guidance, and semantic disentanglement.

\textbf{Global Guidance.}  
We adopt intra- and inter-contrastive loss to align each video moment with the global query representation. 
Specifically, each target moment contains several video clips. We randomly select one as the positive and denote its token as $v^p$. The video clips not from the target moment are negative ones, and their tokens are denoted as $v^n$.
We utilize the weighted pooling \cite{Zhang_Sun_Jing_Zhou_2020} to aggregate the query features and obtain global query representation  $\mathcal{Q}^g \in \mathbb{R}^{1 \times d}$.
The intra-contrastive loss is formulated as follows,
\begin{equation}
\small
\mathcal{L}_\mathrm{intra}^\mathrm{g}=-\frac{1}{B}\sum_{i}^B\log\frac{\exp\left(S^{v_i^p\cdot \mathcal{Q}_i^g} \right)}{\exp\left(S^{v_i^p\cdot \mathcal{Q}_i^g}\right) + \sum_{j}\exp\left(S^{v_j^n \cdot \mathcal{Q}_i^g}\right)},
\end{equation} 
where $S$ represents the cosine similarity, $B$ is the batch size, $i$ represents all indexes in each batch, and $j$ only represents the negative indexes. 
The inter-contrastive loss is formulated as:
\begin{equation}
\mathcal{L}_\mathrm{inter}^\mathrm{g}=-\frac{1}{B}\sum_{i}^B\log\frac{\exp\left(S^{v_i^p\cdot \mathcal{Q}_i^g} \right)}{\sum_{j}^B\exp\left(S^{v_i^p \cdot \mathcal{Q}_j^g}\right)}.
\end{equation}

The intra-contrastive loss enables understanding semantic distribution among different moments within each video. Meanwhile, the inter-contrastive loss measures the correlation between positive video clips and global query representation.

\textbf{Fine-grained Guidance.}
A fine-grained contrastive loss is proposed to perform token-level disentanglement. Firstly, we adopt cross attention between clips $\mathcal{V}^d$ and words $\mathcal{Q}^d$ to reform the original textual information to video-guidance one.
\begin{eqnarray}
\widetilde{\mathcal{Q}^l}=\text{MHA}(\mathcal{V}^d,\mathcal{Q}^d,\mathcal{Q}^d),
\end{eqnarray}
where $\mathcal{V}^d$ is the query and  $\mathcal{Q}^d$ represents key and value. \text{MHA} indicates multi-head cross attention. The video-guidance tokens $\widetilde{\mathcal{Q}^l}=\{w_1,w_2,...,w_T\} $ have the same shape as $\mathcal{V}^d$ and contain the most correlated textual information.


Different from the intra-contrastive loss using global-level representation $\mathcal{Q}^g$, a token-level loss is formulated as:
\begin{equation}
\small
\mathcal{L}^l=-\frac{1}{B}\sum_{i}^B\log\frac{\exp\left(S^{v_i^p\cdot \widetilde{w}_i^l} \right)}{\exp\left(S^{v_i^p\cdot \widetilde{w}_i^l}\right) + \sum_{j}\exp\left(S^{v_j^n \cdot \widetilde{w}_i^l}\right)}.
\end{equation} 
The token-level loss $\mathcal{L}^l$ enables the model to highlight the fine-grained correlation relationship.
Therefore, the total loss for global and fine-grained guidance is formulated as:
\begin{eqnarray}
\mathcal{L}^\mathrm{d}=\lambda^\mathrm{inter}\mathcal{L}_\mathrm{inter}^\mathrm{g}+\lambda^\mathrm{intra}\mathcal{L}_\mathrm{intra}^\mathrm{g}+\lambda^l\mathcal{L}^l.
\end{eqnarray}

\textbf{Semantic Disentanglement.} The video is disentangled into different alignment levels after the two kinds of guidance loss.
A language-guided context encoder~\cite{yang2022improving} $\mathcal{G}$ equipped with two separate multi-head attention is utilized to perform cross-model context aggregation:
\begin{eqnarray}
\mathcal{V}^m = \mathcal{G}(\mathcal{V}^d,\mathcal{Q}^l).
\end{eqnarray}


Then, we calculate relevance score $\mathcal{D} \in \mathbb{R}^T$ QVHighlights~\cite{NEURIPS2021_62e09734} for each video clip as follows:
\begin{eqnarray}
\mathcal{D}_i =S^{v_i^{l},\mathcal{Q}^g}+\frac{1}T  \sum_{j=0}^{T} S^{v_i^{l},\widetilde{w}_j^{l}}.
\end{eqnarray}

The relevance score $\mathcal{D}_i$ is multiplied over each video token $v_i^m$ in $\mathcal{V}^m$, producing disentangled video representations $
\mathcal{V}^q = \{\mathcal{D}_1 v_1^m, \mathcal{D}_2 v_2^m,...,\mathcal{D}_T v_T^m\}$.


\subsection{Context-aware Dynamic Denoising}
\label{subsec:context}
To eliminate the redundant noise and refine the video content to capture the fully aligned events, we propose Context-aware Dynamic Denoising (CDD). The query-relevant spatial-temporal offsets are learned by the proposed context offset mapping to re-aggregate the video context for fine-grained grounding.
Initially, a temporal grid $p = \{p_m\}_{m=1}^{T^{r}}$ is generated as reference points. The grid length $T^{r}$ is down-sampled from $\mathcal{V}^q$ by a factor $r$, implying $T^{r} = T/r$. These reference points are established as linear 1D coordinates ${0,\ldots,(T^{r}-1)}$ according to $T^{r}$ and subsequently normalized within the range $[-1,+1]$. Then input feature $\mathcal{V}^q$ undergoes a linear mapping to $\mathcal{V}^{q'}$ utilizing the parameter $W_q$.

\begin{figure}[ht]
    \centering
    \includegraphics[width=0.35\textwidth]{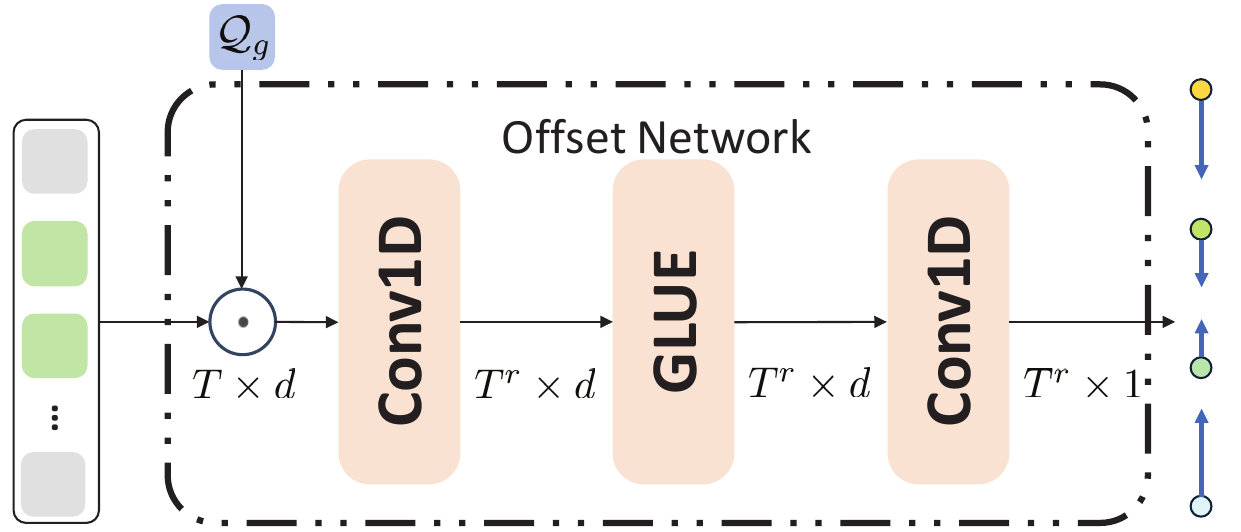}
    \caption{Illustration of offset network.} 
    \label{fig:offset} 
\end{figure}

We introduce the context offset mapping $\Omega_{\text{offset}}$ to generate positional offsets $\Delta p = \{\Delta p_m\}_{m=1}^{T^{r}}$, utilized to determine the displacement of reference 
point locations and eliminate the effects of other noised locations. As depicted in Fig.~\ref{fig:offset}, the context offset mapping $\Omega_{\text{offset}}$ is a lightweight convolutional neural network designed to identify aligned spatio-temporal events within specific video segments, such as highly correlated entities or actions. $\mathcal{V}^{q'}$ is passed into convolutional processing with a kernel size $l$, capturing localized features over a duration of $l$. Subsequently, a GELU activation function followed by a 1x1 convolutional layer is applied to generate offsets for attention points. Crucially, bias in the 1x1 convolutional layer is omitted to prevent unintended offset introduction. A linear-interpolation sampling function $\sigma(\cdot;\cdot)$ is utilized to sample video tokens according learned offsets:
\begin{eqnarray}
\sigma\left(\mathcal{V}^{q'};(p_m+\Delta p_m)\right)=\sum_{t=0}^{T}c(p_m + \Delta p_m,t)\mathcal{V}^{q'}[t,:],
\end{eqnarray}
where $c(p, t)= max(0,1-|p-t|)$. Therefore, the aggregated video representation $\tilde{x}$ can be obtained by capturing less but more effective context through learned offsets:
\begin{align}
&\mathcal{V}^{q'} =\mathcal{V}^{q}W_{q}, \\
&\Delta p =\Omega_{\text{offset}}(\mathcal{V}^{q'} \odot \mathcal{Q}^g), \\
&\tilde{x} =\sigma(\mathcal{V}^{q'};(p_m + \Delta p_m)). 
\end{align}
where $\odot$ denotes Hadamard product. Then we transform $\tilde{x}$ as the key and value and adopt $\mathcal{V}^q$  as the query to perform cross attention, obtaining the final de-noised video representation:
\begin{eqnarray}
&\tilde{\mathcal{V}^k}=\tilde{x}W_{k}, \tilde{\mathcal{V}^v}=\tilde{x}W_{v},\\
&\mathcal{V}^f=\text{MHA}(\mathcal{V}^q,\tilde{\mathcal{V}^k},\tilde{\mathcal{V}_v}),
\end{eqnarray}



\subsection{Cross-modal Content Grounding}
The representations $\mathcal{V}^f$ and $\mathcal{Q}$ are concatenated and fed into the final localizer, producing $\mathcal{V}^{f'}$ and $\mathcal{Q}^{'}$.
The Predict Head comprises two 3-layer convolutional networks $f_a$ and $f_b$ that operate on the output video clip tokens. $f_a$ predicts the confidence level $\hat{p}$ of whether current token $v_t$ is \textbf{fully aligned} to the text query, $\hat{p} = f_a(v^{f'}_t)$. We use a focal loss~\cite{lin2017focal} to optimize strong related confidence:
\begin{equation}
\mathcal{L}^f=-\lambda^\mathrm{f}(1-p_t)^\gamma log(p_t), 
\end{equation}
\begin{equation}
p_t=\{\begin{array}{ll}\hat{p}&\mathrm{if~v_t \in\text{+}}\\
1-\hat{p}&otherwise~, \end{array}
\end{equation}
where $\gamma$ denotes \textit{focusing parameter}.
$f_b$ is responsible for predicting left and right boundary offsets $\hat{d}_t = \left[\hat{d}^{l}_t,\hat{d}^{r}_t\right]$ at each positive video token and $\hat{b}_t = \left[t-\hat{d}^{l}_t,t+\hat{d}^{r}_t\right]$. We utilize a combination of smooth $L1$ loss and generalize IoU loss~\cite{lin2023univtg} to optimize boundary prediction objective,
\begin{equation}\mathcal{L}^{b}=\lambda^{\mathrm{Ll}}\mathcal{L}^{\mathrm{SmoothLl}}\left(\hat{d}_t,d_t\right)+\lambda^{\mathrm{iou}}\mathcal{L}^{\mathrm{iou}}\left(\hat{b}_t,b_t\right),\end{equation}
where $d_t$ and $b_t$ denote the ground truth boundary offsets and bounding boxes, $\lambda_{\mathrm{Ll}}$ and $\lambda_{\mathrm{iou}}$ represents balance parameters.
The over all objective can be formulated as,
\begin{equation}
\mathcal{L}=\frac1T\sum_{i=1}^T\left(\mathcal{L}^\mathrm{d}+\mathcal{L}^\mathrm{f}+\mathcal{L}^\mathrm{b}\right).
\end{equation}

The original outputs is optimized by Hungarian~\cite{carion2020end} matcher and Non-maximum supression~(NMS) with a threshold of 0.7 to obtain the final results.

\begin{table*}[htbp]
\renewcommand\arraystretch{0.8}
  \centering
  \caption{
  The results of Video Moment Retrieval (\textbf{VMR}) and Highlight Detection (\textbf{HD}) on QVHighlights \textit{val} split. The symbol \dag signifies the utilization of the audio modality, while \ddag indicates that the model is pre-trained.}
    \begin{tabular}{cccccccccccc}
    \toprule
    \multirow{3}[6]{*}{Method} & \multicolumn{9}{c}{VMR}  & \multicolumn{2}{c}{HD} \bigstrut\\
    \cmidrule(lr){2-10}\cmidrule(lr){11-12}   & \multicolumn{3}{c}{R1} & \multicolumn{3}{c}{R5} & \multicolumn{3}{c}{mAP} & \multicolumn{2}{c}{$>=$ Very Good} \bigstrut\\
    \cmidrule(lr){2-4}\cmidrule(lr){5-7} \cmidrule(lr){8-10} \cmidrule(lr){11-12}      & @0.3 & @0.5 & @0.7 & @0.3 & @0.5 & @0.7 & @0.5 & @0.75 & avg. & mAP & HIT@1 \bigstrut\\
    \midrule
    CLIP~\cite{radford2021learning} & -   & 16.88 & 5.19 & -   & -   & -   &  18.11  &  7.00 &  7.67  & 31.30   & 61.04 \\
    XML~\cite{Lei_Yu_Berg_Bansal_2020} & -   & 41.83 & 30.35 & -   & -   & -   & 44.63   & 31.73  & 32.14 & 34.49 & 55.25 \\
    UMT\dag~\cite{liu2022umt} & -   & 56.23 & 41.18 & -   & -   & -   & 53.83   & 37.01   & 36.12 & 38.18 & 59.99 \\
    M-DETR~\cite{NEURIPS2021_62e09734} & 67.48   & 53.94 & 34.84 & 91.48  & 75.03   & 47.74   & 54.96  & 31.01   & 32.20 & 35.65 & 55.55 \\
    UniVTG~\cite{lin2023univtg} & 75.16   & 61.56 & 39.42 & 89.55  & 79.35  & 59.94   & 60.42   & 34.76  & 35.47 & 38.20 & 62.65 \\
    QD-DETR\dag~\cite{moon2023query} & 74.77  & 62.90 & 46.77 & 93.48  &  80.90  &  60.65 &  62.66 & 41.51 & 41.24 & 39.49 & 64.13 \\
    QD-DETR\ddag & 75.03  & 63.74 & 47.74 & 92.00  &  81.16  & 59.68  & 62.92  & 42.16  & 41.72 &   39.36  & 65.03 \\
    \textbf{CDNet} (Ours) & \textbf{78.32}   & \textbf{67.74} & \textbf{49.55} & \textbf{93.74} & \textbf{81.42} & \textbf{63.48} & \textbf{63.82} & \textbf{42.30} & \textbf{42.76} & \textbf{39.84} & \textbf{66.52} \bigstrut[b]\\
    \bottomrule   
    \end{tabular}%
  \label{tab:addlabel}%
\end{table*}

\section{Experiments}

\subsection{Implementation Details}
The hidden dimension of CDNet is set to 1024. The down-sampling factor $r$ in CDD is set to 4, while the learning rate is fixed at 1e-4  across all datasets. A drop path rate of 0.1 is employed on the residual connections in MHA layers. Additionally, $\lambda_\mathrm{inter}$, $\lambda_\mathrm{intra}$ and $\lambda_\mathrm{local}$ are set to 0.1, whereas $\lambda_\mathrm{f}$ and $\lambda_\mathrm{b}$ are set to 1. The video clip encoder comprises CLIP~\cite{radford2021learning} using ViT-B/32 and SlowFast~\cite{feichtenhofer2019slowfast} employing ResNet-50, while the text query is encoded using the CLIP text encoder. All experiments are conducted utilizing 4 Tesla V100 GPUs.

\textbf{Datasets.}
We conduct experiments on three representative datasets. \textbf{Charades-STA}~\cite{gao2017tall} comprises intricate daily human activities, encompassing 6768 videos with over 16000 textual queries. \textbf{QVHighlights}~\cite{NEURIPS2021_62e09734} derived from YouTube, encompasses extensive subjects. \textbf{TACoS}~\cite{regneri2013grounding} contains 127 high-quality videos of people performing basic cooking tasks and 2206 aligned textual descriptions.

\textbf{Metrics.}
For VMR task, we follow R$m$@IoU~($m$ denotes top $m$ answers) and AP@IoU or mAP proposed by~\cite{NEURIPS2021_62e09734}. For highlight detection in QVHighligths, we use mAP and HIT@1 when a clip is treated as a true positive if it has the saliency score of ``Very Good".

\subsection{Comparisons with State-of-the-art Models}
To fairly evaluate the performance of our proposed CDNet, we compare it with other state-of-the-art methods on three datasets and represent the results in Tab. \ref{tab:addlabel}, \ref{tab:charades}, and~\ref{tab:tacos}.  As observed, our method outperforms the previous methods without any large-scale pre-training by a significant margin, improving R1@0.3 by at least 3$\%$ at most of the benchmark. Specifically, we attain a +3.29$\%$ improvement in R1@0.3 and a +4.00$\%$ enhancement in R1@0.5 compared to the pre-trained QD-DETR~\cite{moon2023query} in QVHighlights dataset. This suggests that our method acquires sufficiently disentangled and de-noised representations, enabling it to effectively handle videos characterized by intricate scene transitions, which commonly exhibit semantic overlap and ambiguity.

\begin{table}[htbp]
\renewcommand\arraystretch{0.8}
\setlength{\tabcolsep}{1.5pt}
  \centering
  \caption{Results on Charades-STA \textit{test} split.}
    \begin{tabular}{cccccc}
    \toprule
    \multicolumn{1}{c}{Method} & \multicolumn{1}{l}{R1@0.3} & \multicolumn{1}{l}{R1@0.5} & \multicolumn{1}{l}{R1@0.7} & \multicolumn{1}{l}{R5@0.5}  & \multicolumn{1}{l}{mAP} \\
    \midrule
    2D-TAN~\cite{Zhang_Peng_Fu_Luo_2019}& 58.76 & 46.02 & 27.50 & -  & 28.13 \\
    M-DETR~\cite{NEURIPS2021_62e09734} & 65.83 & 52.07 & 30.59 & -  & 37.73 \\
    UMT~\cite{liu2022umt} & -    & 56.23 & 41.18 & -    & 36.12  \\
    UniVTG~\cite{lin2023univtg} & 70.81 & 58.01 & 35.65  & 88.55  & -  \\
    QD-DETR~\cite{moon2023query} & -  & 57.31 & 32.55 &  -   & 39.86 \\
    \textbf{CDNet}(Ours) & \textbf{71.25} & \textbf{58.09} & \textbf{36.53} & \textbf{91.74}  & \textbf{41.06} \\
    \bottomrule
    \end{tabular}%
  \label{tab:charades}%
\end{table}%
\begin{table}[htbp]
\renewcommand\arraystretch{0.8}
  \setlength{\tabcolsep}{4pt}
  \centering
  \caption{Results on TACoS \textit{val} split.}
    \begin{tabular}{cccccc}
    \toprule
    \multicolumn{1}{c}{Method}  & \multicolumn{1}{l}{R1@0.3} & \multicolumn{1}{l}{R1@0.5} & \multicolumn{1}{l}{R1@0.7}  & \multicolumn{1}{l}{mIoU} \\
    \midrule
    2D-TAN~\cite{Zhang_Peng_Fu_Luo_2019} & 40.01 & 27.99 & 12.92  & 27.22 \\
    VSLNet~\cite{Zhang_Sun_Jing_Zhen_Zhou_Goh_2021} & 35.54 & 23.54 & 12.15  & 24.99 \\
    M-DETR~\cite{NEURIPS2021_62e09734} & 37.97 & 24.67 & 11.97  & 25.49 \\
    UniVTG~\cite{lin2023univtg}  & 51.44 & 34.97 & 17.35  & 33.60  \\
    \textbf{CDNet}(Ours)  & \textbf{54.11} & \textbf{35.35} & \textbf{20.34}  & \textbf{33.76} \\
    \bottomrule
    \end{tabular}%
  \label{tab:tacos}%
\end{table}%
\subsection{Ablation Study}

\begin{table}[htbp]
\renewcommand\arraystretch{0.9}
\setlength{\tabcolsep}{2pt}
  \centering
  \caption{Ablation Results. We use a smaller model (hidden dimension is 256) while keeping others the same. $\scriptstyle \times$$\mathcal{D}$ denotes that relevance score $\mathcal{D}$ is multiplied over video tokens.}
    \begin{tabular}{ccccccccc}
    \toprule
    \multicolumn{4}{c}{QSD} & CDD & \multicolumn{2}{c}{VMR} & \multicolumn{2}{c}{HD} \bigstrut\\
\cmidrule(lr){1-4}\cmidrule(lr){6-7}\cmidrule(lr){8-9}   $\mathcal{L}_\mathrm{intra}^\mathrm{g}$  & $\mathcal{L}_\mathrm{inter}^\mathrm{g}$  & $\mathcal{L}^\mathrm{l}$ & \multicolumn{1}{r}{$\scriptstyle \times$$\mathcal{D}$} &     & mAP & R1@0.5 & mAP & HIT@1 \bigstrut\\
    \midrule
        &     &     &     &     & 32.98 & 64.97 & 16.22 & 15.81 \bigstrut[t]\\
    \checkmark &     &     &     &     & 35.01 & 73.13 & 32.18 & 61.35 \\
    \checkmark & \checkmark &     &     &     & 35.38 & 74.52 & 34.65 & 62.10 \\
    \checkmark & \checkmark & \checkmark &     &     & 36.23 & 74.88 & 36.64 & 62.87 \\
    \checkmark & \checkmark & \checkmark & \checkmark &     & 37.34 & 75.31 & 37.24 & 63.55 \\
    \checkmark & \checkmark & \checkmark & \checkmark & \checkmark & 39.98 & 76.92 & 38.85 & 64.71 \bigstrut[b]\\
    \bottomrule
    \end{tabular}%
  \label{tab:ablation}%
\end{table}%

\textbf{Effects of QSD.}
We investigate the impact of various disentangling losses within the proposed QSD and present the results in Tab.~\ref{tab:ablation}. Through the utilization of global guidance, involving both intra and inter-contrastive losses, the model significantly enhances its capacity to comprehend aligned video content. Additionally, the integration of fine-grained guidance contributes to further enhancements in performance. Simultaneously, we measure the accuracy of the proposed relevance score by employing highlight detection (HD) as the metric. The results indicate that QSD effectively captures disentangled semantic distributions across varied moments. Moreover, leveraging relevance scores as adjusters facilitates the extraction of disentangled video representations, leading to substantial performance improvements.



\begin{figure}[thbp!]
\begin{minipage}[t]{0.49\linewidth}
  \centering
  \includegraphics[width=1\textwidth]{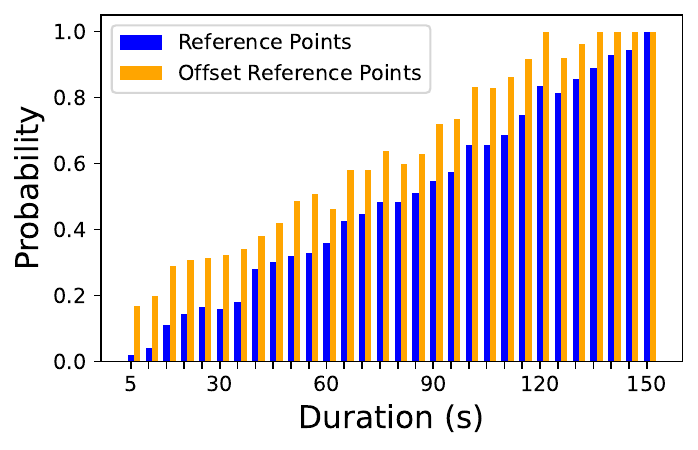}
  \subcaption{}
\end{minipage}
\begin{minipage}[t]{0.49\linewidth}
  \centering
  \includegraphics[width=1\textwidth]{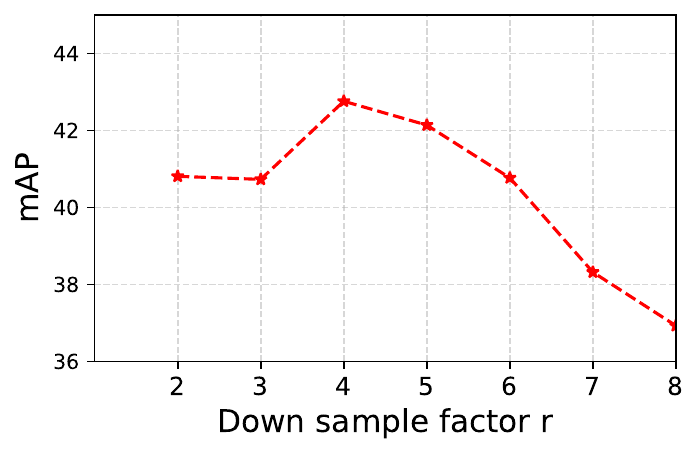}
  \subcaption{}
\end{minipage}
\caption{(a) Probability of reference point belongs to the target moment. (b) Effects of down-sample factor $r$. } 
\label{fig:vis1}
\end{figure}

\textbf{Effecs of CDD.}
To demonstrate the effectiveness of integrating CCD and QSD, we present results of CDNet in the lower section of Tab.\ref{tab:ablation}, showcasing a notable 2.64\% improvement in mAP. Moreover, in Fig.\ref{fig:vis1} (a), we illustrate the impact of linear interpolation sampling. The x-axis represents the varying duration of moments within QVHighlights, while the y-axis depicts the probability of sampled points belonging to specific moments. The reference offsets, computed by $\Omega_{offset}$, elevate these points' probability of aligning with the target moment by an average of 18.3\%. Subsequently, Fig.~\ref{fig:vis1} (b) depicts the alteration in the down-sample factor. A higher down-sample factor indicates sparser sampled distributions, causing a significant loss of video content and subsequent performance degradation. Conversely, a lower down-sample rate introduces noisy content due to the higher number of points. Consequently, we adopt $r=4$ as our setting, yielding the optimal mAP of 42.76\%.


\subsection{Qualitative Analysis}

\begin{figure}[ht]
    \centering
    \includegraphics[width=0.48\textwidth]{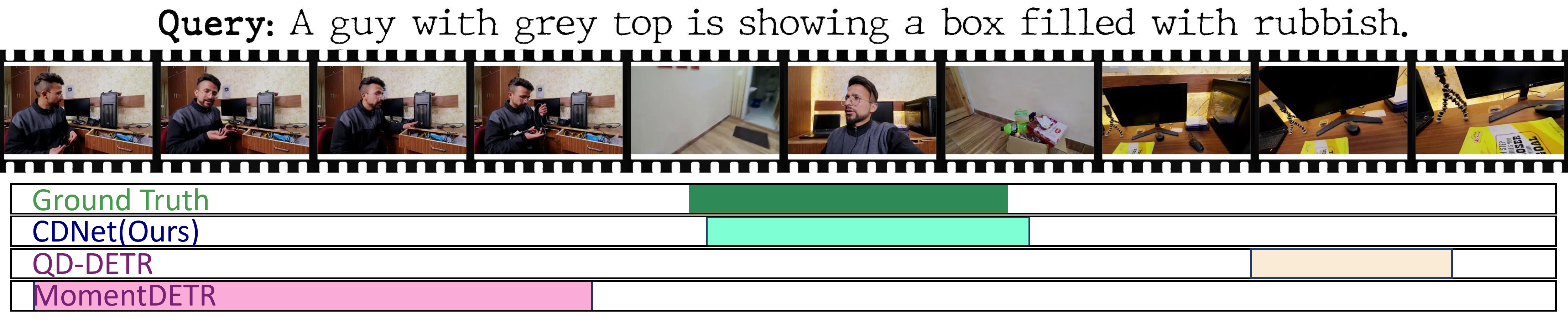}
    \includegraphics[width=0.48\textwidth]{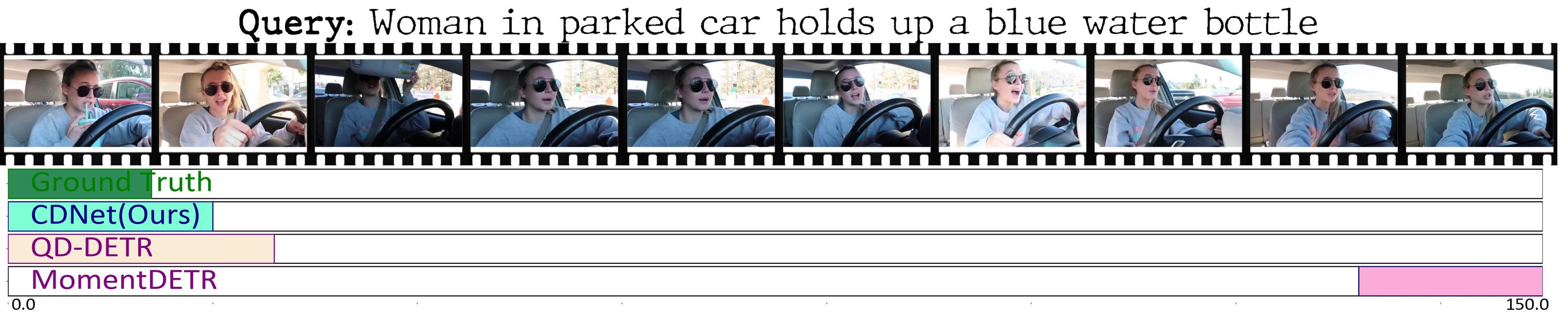}
    \caption{Visualized comparisons of earlier methods~\cite{NEURIPS2021_62e09734,moon2023query} and CDNet. The top green bar represents \textit{ground truth}.} 
    \label{fig:vis} 
\end{figure}

We carefully choose representative samples and visually compare CDNet and earlier methods (Moment-DETR~\cite{NEURIPS2021_62e09734} and QD-DETR~\cite{moon2023query}). Examining the upper portion of Fig.\ref{fig:vis}, we find that CDNet effectively captures the complete semantic context of textual queries and accurately identifies specific moments. Conversely, QD-DETR and M-DETR merely align with partial textual semantics, such as 'person' or 'clutter', leading to unsuccessful video moment retrieval. Additionally, in scenarios where a video lacks changes in actions or scenes, precise comprehension of nuanced contexts becomes essential for the model, as shown in the bottom section of Fig.\ref{fig:vis}. 
Unlike QD-DETR, CDNet establishes boundaries more aligned with the semantic essence of the target moment, while M-DETR produces inaccurate interpretations.

\section{Conclusion}
This work introduces the Contexting Denoising Network (CDNet) tailored for the Video Moment Retrieval task. Comprising the Query-guided Semantic Disentangling (QSD) and the Context-aware Dynamic Denoising (CDD), CDNet is specifically designed based on two key priors in the VMR task: (1) the partial alignment of annotated text and video, (2) the existence of noise and redundancy in the temporal context of video moments. Extensive evaluations validate the effectiveness of the proposed modules. 

\bibliographystyle{ieeetr}
\bibliography{main}
\end{document}